\documentclass[runningheads]{llncs}
\usepackage{graphicx}
\usepackage{multirow}
\begin{document}
\title{KATSum: Knowledge-aware Abstractive Text Summarization}
%
%
 \author{Guan Wang \inst{1} \and
         Weihua Li \inst{1} \and
         Edmund Lai \inst{1} \and 
         Jianhua Jiang \inst{2}
       }
 \authorrunning{Wang et al.}
%
 \institute{Auckland University of Technology, Auckland, New Zealand \\
 \email{guan.wang@autuni.ac.nz \{weihua.li, edmund.lai\}@aut.ac.nz} \and
 Jilin Province Key Laboratory of Fintech, Jilin University of Finance and Economics, China \\
 		\email{jjh@jlufe.edu.cn}
 }
\maketitle              
\begin{abstract}
Text Summarization is recognised as one of the NLP downstream tasks and it has been extensively investigated in recent years. It can assist people with perceiving the information rapidly from the Internet, including news articles, social posts, videos, etc. Most existing research works attempt to develop summarization models to produce a better output. However, advent limitations of most existing models emerge, including unfaithfulness and factual errors. In this paper, we propose a novel model, named as Knowledge-aware Abstractive Text Summarization, which leverages the advantages offered by Knowledge Graph to enhance the standard Seq2Seq model. On top of that, the Knowledge Graph triplets are extracted from the source text and utilised to provide keywords with relational information, producing coherent and factually errorless summaries. We conduct extensive experiments by using real-world data sets. The results reveal that the proposed framework can effectively utilise the information from Knowledge Graph and significantly reduce the factual errors in the summary.

\end{abstract}

\section{Introduction}
Text Summarization aims to produce short and brief texts for long documents while keeping the core information. It can be classified into two kinds of methods, i.e., \textbf{(1)} {\itshape extractive methods} \cite{zhong2020extractive,xu2019discourse}, which formulate the problem as a classification task or a text ranking task, aiming at identifying the most meaningful phrases, sentences and combining them together without modification; \textbf{(2)} {\itshape abstractive methods} \cite{dou2020gsum,see2017get}, where the summary is composed of the words or phrases that are generated without constraints. In other words, the abstractive models can generate novel words other than using the existing ones in the source text. On top of that, some studies attempt to fuse both methods, namely, identifying the key information and generating coherent text by using the extractive method and abstractive method, respectively \cite{zhang2020pegasus,liu2019text}. Such approaches can also help mitigate the non-fluency issue from extractive methods and address the intractable problem resulting from the abstractive methods.

Based on the above comparisons, it is evident that the extractive methods are capable of producing a summary that closely follows the grammar rules and presents the facts since the summary consists of text chunks copied over from the source text without any modification. In other words, the selected text chunks are supposed to be concatenated to produce a short passage as a summary. However, by leveraging such an approach, the summary appears not coherent since two sentences in a summary may be far from each other in the source text. Compared with the extractive methods, the abstractive methods can produce cohesive text. However, they suffer from two major issues, i.e., unfaithfulness and factual inconsistency. The former indicates that the content of the generated summary is far from the main idea of the source text, failing to retain the salient information. The latter refers to the issue that the summarization generator produces incorrect facts from the source text due to its unconstrained nature. Studies have shown that abstractive generator tends to distort the facts from the source text \cite{kryscinski2019evaluating,maynez2020faithfulness}, and 30\% summaries generated through abstractive methods suffer from the factual inconsistency \cite{cao2018faithful}. In this sense, it is vital to mitigate this issue and avoid producing misleading texts.

Many research works employ keywords in the summarization models, ensuring that the key information from the source text is incorporated. For instance, Li et al. develop an extractor to extract keywords as guidance, which can prevent the model from losing key information \cite{li2018guiding}. Zhou et al. propose a selective gate network to select keywords from the source text \cite{zhou2017selective}. See et al. utilise the pointer mechanism and propose a pointer-generator model to identify the keywords \cite{see2017get}. However, the relations, representing the semantic relationships between the keywords, are neglected, leading to unfaithfulness and factual errors in the summary. 

Knowledge Graph has been widely acknowledged as a suitable tool for modelling connected data, where the nodes present entities and links are modelled as corresponding semantic relationships. In this paper, we aim to propose a novel text summarization framework by leveraging the advantages offered by Knowledge Graph. The proposed framework is capable of producing coherent text and emphasising the facts extracted from the source text, where the Knowledge Graph guides the summarization generation process. Specifically, we utilise the Knowledge Graph triplets, in the form of $\{head, relation, tail\}$, to incorporate keywords with their relationships in the process of text summarization. Different from the existing keyword extraction methods, we present the summarization process as a pipeline, starting from Knowledge Graph construction. Then, the extracted Knowledge Graph will be mapped into a lower-dimensional embedding space, in which each triplet embedding is supposed to be fed into a trained classifier, checking if this triplet should be included in the summary or not. To evaluate the proposed framework, we conduct experiments on CNN/Daily Mail data set and calculate the ROUGE scores. The experimental results reveal that our model can yield a better performance by comparing against the baseline models.

The rest of the paper is organised as follow. In Section 2, the related works and limitations are articulated. In Section 3, we elaborate on the proposed novel text summarization model. Sections 4 and 5 introduce the experimental set-up and the experimental results, respectively. Finally, the paper has been concluded in Section 6. 

\section{Related works}
\subsection{Abstractive Summarization}
Abstractive summarization appears more challenging than extractive summarization since it requires the ability to understand the article and generate summaries as humans do.  
Many research works have been dedicated to improving the ability of paraphrasing the source text based on abstractive generators. For example, Rush et al. apply the neural encoder-decoder architecture to text summarization and discuss the possible encoder choices \cite{rush2015neural}. Based on this work, many other researchers conduct their studies on text summarization by enhancing the capability of encoding and decoding, solving problems, i.e., out-of-vocabulary and repetition issues. For example, See et al. utilise the pointer and coverage mechanisms to address the out-of-vocabulary and repetition issues. Meanwhile, the pointer-generator network attempts to precisely reproduce the fact information of the source text \cite{see2017get}. Gehrmann et al. address the content selection issue and propose a two-step process \cite{gehrmann2018bottom}. The first step is content selection, considered as a token-level sequence tagging task. The second step is bottom-up copy attention, restricting the attention to over-determine the selection of source text fragments. 

\subsection{Fact-aware Summarization}

In recent years, fact-aware summarization has attracted attention from researchers. It has been showing the challenges of generating factually correct summaries. Previous studies have proved that abstractive summarization generally suffered from hallucinating phenomena \cite{kryscinski2019evaluating}. Specifically, the text summarization generator sometimes tends to manipulate the facts from the source text, and approximate 30\% of summaries from state-of-the-art models experience the factual inconsistency issue \cite{cao2018faithful}.

In order to alleviate this phenomenon, Cao et al. propose a fact-aware neural abstractive summarization to encode the extracted facts into the encoder along with the source text \cite{cao2018faithful}. Kryscinski et al. propose a novel approach to verify the factual consistency and identify the conflicts between the source text and summary \cite{kryscinski2019evaluating}. Li et al. argue that a summary is meant to be semantically entailed by the source text, treating the fact-aware summarization as an entailment-aware process \cite{li2018ensure}. Based on this argument, they propose an entailment-aware summarization system to incorporate entailment knowledge into the summarization models, which can alleviate the factual inconsistency issue. Zhu et al. utilise Knowledge Graph to integrate factual information into the summarization process \cite{zhu2020boosting}. However, other unrelated information is also involved in the text summarization process, which significantly impacts the quality of the generated summary. 

\subsection{Keywords Extraction}
Classic abstraction models learn the distribution over the input texts and generate the token-level or sentence-level representations. These vectors are fed into the decoder that produces a summary with the trained model parameters that can maximize the output likelihood. However, due to the unconstrained nature of neural-based language models, the output sometimes is difficult to control. Thus, many researchers attempt to adopt constraints during the text summary-generating process. For example, Zhou et al. propose a selective gate network to extract key points, leading the summary to be more focused on these extracted keywords \cite{zhou2017selective}. Ercan et al. leverage WordNet in the lexical chain-building algorithm, which can discover the relations between two word senses \cite{ercan2007using}. Building lexical chains turns out to be a very time-consuming work since it relies on exhaustive algorithms. Litvak et al. \cite{litvak2008graph} propose a graph-based method of both supervised and unsupervised approaches. In the supervised approach, the novel method first converts the documents into word graphs and trains a classifier to check if the corresponding word is included in the document extractive summary. In the unsupervised approach HITS algorithm \cite{kleinberg1998authoritative} is utilised to calculate the weight of each node and select the top ones.

The aforementioned works have achieved promising results. However, the limitations are still presented. Classical abstractive summarization models do not consider the impact of additional reference from the source text but generate summaries only relying on the hidden states of the inputs. Fact-aware models apply a fact-aware approach to guide the generating process, but unrelated information is also used as guidance. Such unrelated information may distract the summarization generator from producing a better summary. Meanwhile, approaches with keyword extraction neglect the relation of these keywords, and the generator may manipulate the facts. 

In this paper, we propose a novel Knowledge-aware Abstractive Text Summarization (KATSum) model, which can address the limitations mentioned above. We utilise the Knowledge Graph to extract triplets, eliminate the noise with a trained classifier and identify useful triplets as auxiliaries to guide the text summarization generation process.

\section{Knowledge-aware Abstractive Text Summarization}
In this section, we introduce the proposed novel text summarization model, called Knowledge-aware Abstractive Text Summarization (KATSum) model, by explaining the architecture and triple extraction process. 

\begin{figure}[t]
	\includegraphics[width=1.0\textwidth]{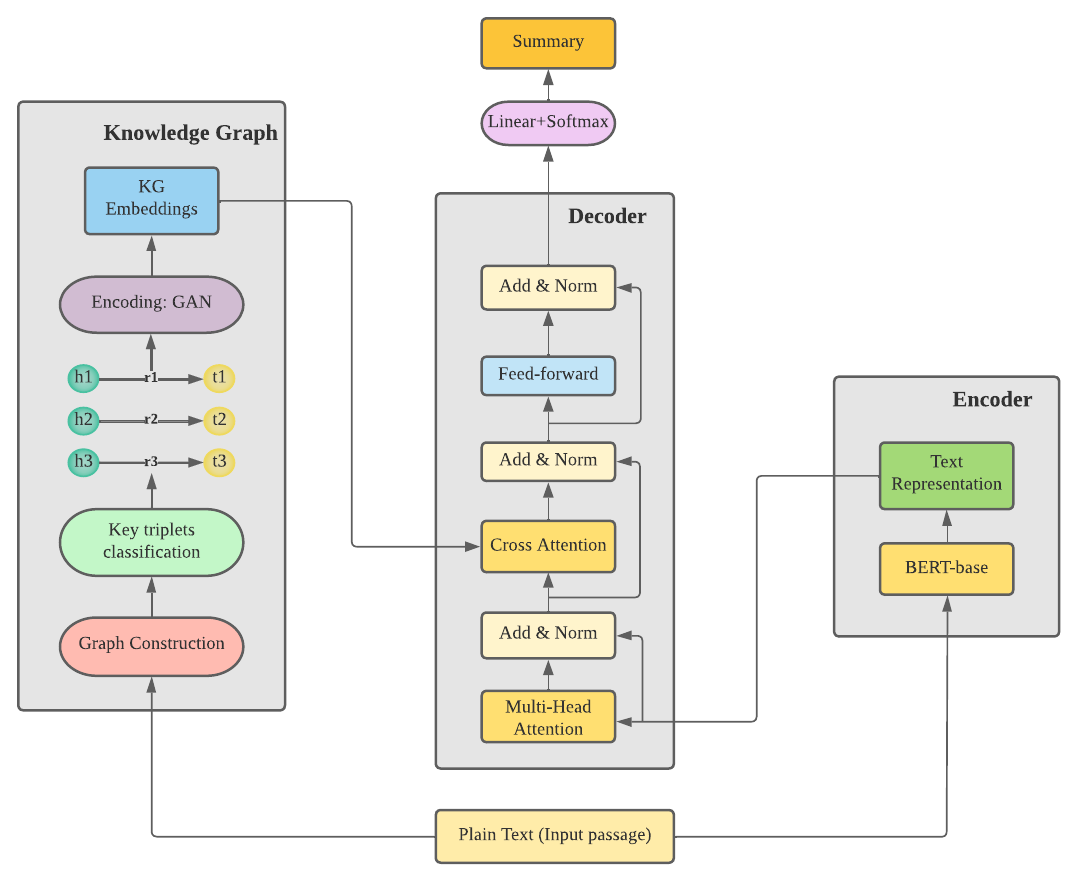}
	\caption{The Architecture of KATSum} \label{fig1}
\end{figure}

The architecture of KATSum has been presented in Figure~\ref{fig1}. The proposed KATSum model consists of a knowledge-aware encoder encoding the article and identifying key information, and a decoder generating texts. The knowledge-aware encoder incorporates two pipelines, i.e., the classic encoder pipeline and the knowledge-graph pipeline. The former is composed of a pre-trained BERT model \cite{devlin2018bert}, which produces the hidden states used as part of the decoder input. The latter extracts key triplets from the source texts and maps them into an embedding vector space. Before feeding into the decoder, the outputs from the BERT encoder and knowledge graph are fused as the decoder input.

\begin{figure}[t]
	\includegraphics[width=1.0\textwidth]{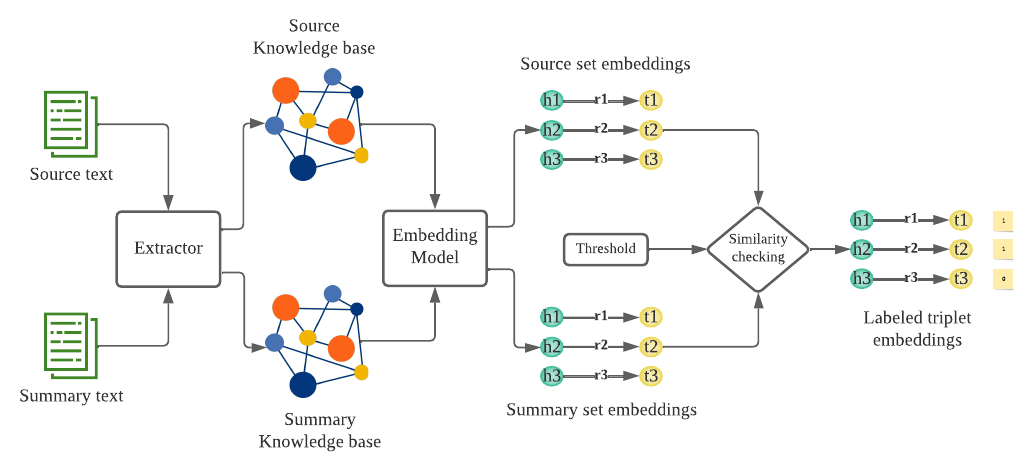}
	\caption{The process of extracting triplets from source text.} \label{fig2}
\end{figure}


On top of that, we employ a Knowledge Graph embedding classifier to identify the knowledge included in the summary from the source text. Triplets are extracted from both source texts and summaries to train this classifier, and presented as source triplet set and summary triplet set, respectively. In order to identify the key triplets, TransE \cite{bordes2013translating} has been adopted to transform the extracted Knowledge Graph into a low-dimensional embedding space so that the semantic-level similarity can be compared. Each triplet from the source set is labelled as 1 if a similar triplet is found in the summary triplet set, and 0 otherwise. We train the Knowledge Graph embedding classifier to identify the key information from the source text by using the labelled data. The information from the source text can be extracted in the form of a triplet, i.e., $\{head, relation, tail\}$. The Sigmoid classifier is formulated in Equation \ref{eq:sigmoid}. 

\begin{equation}
	\label{eq:sigmoid}
	\hat{y}_i = \sigma(We_i + b), 
\end{equation}

\noindent where $e_i$ denotes the vector of Triplet $t_i$. The identified triplet embedding is fused into the baseline model from where we get the Vector $v_i$.

The decoder of KATSum derives from the Transformer Decoder with 6 identical layers. Inspired by \cite{liu2019text}, we follow the training strategy using two Adam optimizers respectively with different warm-up schedules for the learning rate:
\begin{equation}
	lr_e = \widetilde{lr}_e\cdot min(step^{-0.5}, step\cdot warmup^{-1.5}_e)
\end{equation}
\begin{equation}
	lr_d = \widetilde{lr}_d\cdot min(step^{-0.5}, step\cdot warmup^{-1.5}_d)
\end{equation}

\section{Experimental Setup}

In this section, we first formulate the text summarization problem. Then, the data sets and baselines used in the experiments are introduced. Finally, we explain the parameter selections.

\subsection{Text Summarization Problem} 
Given an input sequence, $X = \{x_1, x_2, x_3,..., x_n\}$, where $n$ refers to the sequence length. The objective is to obtain a text summary, $Y = \{y_1, y_2, y_3, ..., y_m\}$,  $m \leq n$, where $m$ refers to the summary length. 


\renewcommand{\arraystretch}{1.5} 
\begin{table}[t]
	\caption{The statistics of two data sets}
	\label{tab1}
	\centering
	\resizebox{\textwidth}{!}{
	\begin{tabular}{ccccccc}
		\hline
		\multirow{2}{*}{\textbf{Data Set}} & \multirow{2}{*}{\textbf{Data Type}} & \multicolumn{3}{c}{\textbf{Sample Pairs}}                  & \multicolumn{2}{c}{\textbf{Avg Tokens}} \\ \cline{3-7} 
		&                                & \textbf{Training} & \textbf{Validating} & \textbf{Testing} & \textbf{Source Text} & \textbf{Summary} \\ \hline
		CNN/Daily Mail                             & News                           & 287,226           & 13,368               & 11,490            & 781                & 56             \\
		XSum                               & News                           & 204,045            & 11,332               & 11,334            & 431                & 23             \\ \hline
	\end{tabular}
	}
\end{table}
	
	\subsection{Experimental Data Sets}
	Two real-world data sets have been utilised in the experiments, i.e., CNN/Daily Mail \cite{Hermann2015,see2017get} and XSum \cite{Narayan2018}. The statistics of both data sets are listed in Table~\ref{tab1}. 
	
	\begin{itemize}
		\item {\bfseries CNN/Daily Mail } is a widely used data set in many research works especially has been used for evaluating text summarization. It consists of news articles (781 tokens on average) from CNN and Daily Mail websites paired with multi-sentence summaries (3.75 sentences or 56 tokens on average). It in total contains 287,226 (92\%) training pairs, 13,368 (4.3\%) validation pairs and 11,490 (3.7\%) test pairs. 
		
		\item {\bfseries XSum} is a data set of articles (431 tokens on average) and one-sentence summaries (23 tokens on average) collected from BBC. The official random split includes 204,045 (90\%), 11,332 (5\%) and 11,334 (5\%) documents in training, validation and test sets, respectively.
	\end{itemize}

	\subsection{Evaluation Metrics}
	ROUGE \cite{lin2004rouge} is selected as the evaluation metric for measuring the quality of the generated summaries. ROUGE\_1 and ROUGE\_2 refer to the overlap of uni-gram and bi-gram between the source text and the generated summary, respectively. ROUGE\_L describes the longest common sub-sequence. ROUGE is one of the most widely used evaluation metrics for text summarization \cite{lin2004rouge}. It focuses on token-level matching but ignores the semantic-level matching between the summary and the source text, which results in the inability to identify factual errors. 
	
	\subsection{Baselines}
	To evaluate KATSum, we utilise the existing models with encoder-decoder architecture as the baselines. Specifically, the transformer-based pre-trained language models, e.g., BERT \cite{devlin2018bert}, are adopted as the encoder. For the decoder, we select the original transformer decoder with six identical layers \cite{klein2018opennmt}. To accommodate the BERT input, three types of embeddings, i.e., token embeddings, segment embeddings and position embeddings, are fused as one input vector before being fed into the encoder.
	
	There are many other pre-trained language models. Different combinations of these models, e.g., BERT as encoder and GPT2 as decoder, can yield a different performance. Therefore, we select different combinations of several pre-trained language models as the baselines. In this paper, the optional encoders include BERT and XLNet \cite{yang2019xlnet}. As for the decoder, the original transformer decoder is applied. 
	
	\subsection{Parameters Selection}
	As the encoder of the KATSum backbone, we leverage "bert-base-uncased" implemented by Hugging Face\footnote{https://huggingface.co/}. While the decoder possesses 768 hidden units, and the hidden size of the feed-forward layer is 2,048. 
	
	As the decoder is transformer-based, we train the model by using the same strategy as \cite{liu2019text} since the transformer decoder needs to be trained from scratch while the BERT encoder is well pre-trained. We fine-tune the encoder and train the decoder separately with different schedules. For encoder, we use $\beta_1 = 0.9$ for Adam optimizer, $\hat{lr}_e = 2e^-3$ as the initial learning rate, $warmup_e = 20,000$. For decoder, we use $\beta_2 = 0.999$ for the Adam optimizer, $\hat{lr}_d = 0.1$ as the initial learning rate, $warmup_d = 10,000$. The parameter settings are given according to the empirical values from the existing studies. 
	
	To label the Knowledge Graph triplets, we also conduct experiments on the similarity threshold. Given the choices, i.e., $threshold \in \{0.5, 0.8, 0.9 \}$, the model yields the best performance when $threshold = 0.8$. To initiate the training, we first train the classifier based on the labelled data for 5 epochs, having 10,000 steps in each epoch. Then, we train the model by using the original data for 200,000 steps. Next, we evaluate and save checkpoints every 2,500 steps. As the GPU memory limits the batch size, we employ the gradient accumulation and calculate the gradient every 5 steps. All the models are trained on 1 Tesla P100 GPU with a memory of 16GB.
	
	\section{Experimental Results}
	
	Two experiments have been conducted. The first experiment aims to evaluate KATSum by comparing it against the other baselines using two data sets. In the second experiment, we conduct an ablation study to demonstrate the effectiveness of the Knowledge Graph module. 
	
	\subsection{Experiment 1}
	
	In this experiment, we calculate ROUGE scores on both CNN/Daily Mail and XSum to compare the performance. The experimental results are illustrated in Tables \ref{tab2} and \ref{tab3}. Two baseline models are implemented. One adopts BERT as the encoder, while the other employs XLNet as the encoder. 
	
	The experimental results explicitly show that our model can remarkably outperform the baselines with the assistance of Knowledge Graph, especially when having XLNet as the encoder. The score of our model with XLNet is more than 20\% higher than the baseline on CNN/Daily Mail. While by using XSum, KATSum with XLNet also outperforms other models, especially the ROUGE\_L scores.

\renewcommand{\arraystretch}{1.5} 
\begin{table}[t]
\caption{ROUGE results on CNN/Daily Mail over different models}
\label{tab2}
\centering

	\begin{tabular}{lccc}
		\hline
		\multicolumn{1}{c}{\textbf{Model}}                       & \textbf{ROUGE\_1} & \textbf{ROUGE\_2} & \textbf{ROUGE\_L} \\ \hline
		Baseline (BERT - Transformer decoder) & 32.5             & 14.2             & 29.8             \\
		Baseline (XLNet - Transformer decoder)           & 34.2             & 15.6             & 31.5             \\ \hline
		KATSum (BERT)                          & 41.4             & 20.3             & 39.5             \\
		KATSum (XLNet)                         & \textbf{42.8}    & \textbf{20.7}    & \textbf{40.2}    \\ \hline
	\end{tabular}

\end{table}

\renewcommand{\arraystretch}{1.5} 
\begin{table}[t]
	\centering
	\caption{ROUGE results on XSum over different models}
	\label{tab3}
	\begin{tabular}{lccc}
		\hline
		\multicolumn{1}{c}{\textbf{Model}}                       & \textbf{ROUGE\_1} & \textbf{ROUGE\_2} & \textbf{ROUGE\_L} \\ \hline
		Baseline (BERT - Transformer decoder) & 33.7              & 14.6              & 27.2              \\
		Baseline (XLNet - Transformer decoder)           & 34.1              & 14.3              & 28.4              \\ \hline
		KATSum (BERT)                          & 44.8              & 22.4              & 38.1              \\
		KATSum (XLNet)                         & \textbf{45.6}     & \textbf{22.9}     & \textbf{38.4}     \\ \hline
	\end{tabular}
\end{table}
	
	Different pre-trained language models are employed as our backbone encoders. We conduct analysis and discuss the results. Intuitively, XLNet is more suitable for working with Knowledge Graph since XLNet can encode the entire article without length restrictions \cite{yang2019xlnet}. Specifically, BERT only uses the first 512 tokens of the source text, and the rest are not encoded. In contrast, the Knowledge Graph can offer key information extracted from the entire source text. In this case, the decoder may not effectively utilise the key information. XLNet has been proved to be effective on increasing the performance of a summarizer \cite{guan2020survey}. Thus, the XLNet has also been applied as the encoder of KATSum. 

\subsection{Experiment 2}	In this experiment, we conduct an ablation study to demonstrate the importance of the Knowledge Graph component of KATSum. Specifically, we first remove the classification process from the Knowledge Graph component and directly use the knowledge graph embeddings from the graph neural network, and observe the impact on the final results. And then we remove the whole Knowledge Graph component from the model and compare the results with the complete model with the Knowledge Graph component and classification process. We use 1/3 of the entire CNN/Daily Mail data set and conduct experiments using the Bert-base KATSum model. 
	
As can be observed from Table ~{\ref{tab4}}, the model with the classification process (removing noisy information) can yield a better performance. From Table ~{\ref{tab5}}, we can observe that the Knowledge Graph component brings contributions to the text summarization, and it can help the model generate better summaries. The Knowledge Graph constructed based on the source text incorporate the complete information. Whereas some information appears not as important as the others, and it may distract the generator from identifying the main idea.

\renewcommand{\arraystretch}{1.5} 
\begin{table}[t]
	\caption{ROUGE results under the condition of with and without the classification process. The model is BERT-based, where 1/3 of the data set is utilised to conduct the experiment.}
	\label{tab4}
	\centering
	\begin{tabular}{lccc}
		\hline
		\multicolumn{1}{c}{\textbf{Model}}                 & \textbf{ROUGE\_1} & \textbf{ROUGE\_2} & \textbf{ROUGE\_L} \\ \hline
		KATSum (with classification)    & 31.7              & 11.3              & 28.8              \\
		KATSum (without classification) & 29.5              & 10.6              & 27.4              \\ \hline
	\end{tabular}
\end{table}

\begin{table}[t]
	\caption{ROUGE results under the condition of with and without the Knowledge Graph component. The model is BERT-based, where 1/3 of the data set is utilised to conduct the experiment.}
	\label{tab5}
	\centering
	\begin{tabular}{lccc}
		\hline
		\multicolumn{1}{c}{\textbf{Model}}                  & \textbf{ROUGE\_1} & \textbf{ROUGE\_2} & \textbf{ROUGE\_L} \\ \hline
		KATSum (with KG component)    & 31.8              & 11.2              & 28.8              \\
		KATSum (without KG component) & 26.7              & 9.3              & 25.2              \\ \hline
	\end{tabular}
\end{table}

\section{Conclusion and Future Work}
	
In this paper, we present a novel knowledge-aware text summarization model, called KATSum. The novel model employs Knowledge Graph to improve the quality of the summaries in terms of ROUGE scores. With the knowledge-aware encoder, the input text will be processed by the pre-trained language model and converted into Knowledge Graph embeddings. Such features can help to address unfaithfulness and factual inconsistency. We conducted two experiments to evaluate KATSum using two real-world data sets. The experimental results have shown that the KATSum significantly outperforms the baselines with pre-trained models.
	
In the future, we will try to calibrate the summaries utilising external resources. As for the Knowledge Graph construction, we plan to investigate the impact of different triplets quality on text summarization.

\newpage

\bibliographystyle{splncs04}
\bibliography{ref}

\end{document}